\documentclass[conference]{IEEEtran}
\IEEEoverridecommandlockouts
\usepackage{cite}
\usepackage{amsmath,amssymb,amsfonts}
\usepackage{algorithmic}
\usepackage{graphicx}
\usepackage{textcomp}
\usepackage{subcaption}
\usepackage{booktabs}
\usepackage{xcolor}
\def\BibTeX{{\rm B\kern-.05em{\sc i\kern-.025em b}\kern-.08em
    T\kern-.1667em\lower.7ex\hbox{E}\kern-.125emX}}
\begin{document}

\title{AttnBoost: Retail Supply Chain Sales Insights via Gradient Boosting Perspective\\
\thanks{$^*$ Equal Contribution; $^{\dag}$ Corresponding Author}
}
\author{
\IEEEauthorblockN{Yadi Liu$^{1,*}$, Xiaoli Ma$^{2,*}$, Muxin Ge$^{3}$, Zeyu Han$^{3,4}$, Jingxi Qiu$^{4}$, Ye Aung Moe$^{5}$,\\ Yilan Shen$^{6}$, Wenbin Wei$^{1}$, Cheng Huang$^{3,\dag}$}
\IEEEauthorblockA{\textit{$^{1}$Nanyang Technological University, $^{2}$Washington State University, $^{3}$Southern Methodist University,} \\
\textit{$^{4}$Georgetown University, $^{5}$University of Nebraska-Lincoln, $^{6}$China Construction Bank}\\
yadi001@ntu.edu.sg, xiaoli.ma@wsu.edu}
}

\maketitle

\begin{abstract}
Forecasting product demand in retail supply chains presents a complex challenge due to noisy, heterogeneous features and rapidly shifting consumer behavior. While traditional gradient boosting decision trees (GBDT) offer strong predictive performance on structured data, they often lack adaptive mechanisms to identify and emphasize the most relevant features under changing conditions. In this work, we propose AttnBoost, an interpretable learning framework that integrates feature-level attention into the boosting process to enhance both predictive accuracy and explainability. Specifically, the model dynamically adjusts feature importance during each boosting round via a lightweight attention mechanism, allowing it to focus on high-impact variables such as promotions, pricing, and seasonal trends. We evaluate AttnBoost on a large-scale retail sales dataset and demonstrate that it outperforms standard machine learning and deep tabular models, while also providing actionable insights for supply chain managers. An ablation study confirms the utility of the attention module in mitigating overfitting and improving interpretability. Our results suggest that attention-guided boosting represents a promising direction for interpretable and scalable AI in real-world forecasting applications.

\end{abstract}

\begin{IEEEkeywords}
Supply Chain Analysis, Attention Mechanism, EXtreme Gradient Boosting, Profit Forecast and Classification
\end{IEEEkeywords}

\section{Introduction}

Accurate demand forecasting is a cornerstone of effective retail supply chain management, enabling businesses to optimize inventory control, replenishment strategies, and promotional planning \cite{bg1,bg3}. Yet, this task remains exceptionally challenging due to retail data's noisy, high-dimensional nature, compounded by volatile demand drivers such as shifting consumer preferences, seasonal fluctuations, competitive pricing dynamics, and promotional campaigns  \cite{bg2,bg4}. While traditional statistical methods offer interpretability, they often fail to capture the complex nonlinear relationships inherent in these factors. As retailers increasingly adopt artificial intelligence for decision support, there is a growing demand for models that balance predictive accuracy with practical interpretability \cite{in}, ensuring robust forecasts and actionable insights for supply chain stakeholders. 

In recent years, GBDT \cite{dt}, such as XGBoost \cite{xgb} and LightGBM \cite{gbm}, have become the de facto standard for predictive modeling in structured data domains. These models offer several distinct advantages: robust performance on tabular data, inherent resistance to overfitting, and straightforward deployment in real-world systems. The GBDT framework operates by iteratively constructing ensembles of weak learners through a stage-wise, greedy minimization of the loss function \cite{bg5}. Despite their success, conventional GBDT models suffer from a fundamental limitation in their static treatment of feature importance. Once a feature is selected for splitting, its relative influence remains fixed throughout all subsequent trees. More critically, these models lack the adaptive capability to dynamically reweight features in response to evolving data distributions or context-specific patterns. This rigidity proves particularly problematic in non-stationary environments such as retail sales forecasting \cite{bg6}, where feature relevance often changes rapidly.

The concept of attention \cite{attn}, initially developed in the context of neural sequence modeling, has proven remarkably effective at dynamically focusing on relevant input components. Recent adaptations to tabular data, as seen in TabNet \cite{tab} and FT-Transformer \cite{ft}, demonstrate that attention mechanisms can simultaneously enhance both model performance and interpretability by adaptively weighting features according to context.

However, deep attention-based models often require extensive training resources and lack transparency when applied to structured datasets.  In this work, we propose integrating a lightweight attention mechanism directly into the GBDT framework, creating an adaptive yet interpretable learning system. By embedding attention, guided feature reweighting into each boosting round, the model can dynamically focus on high-impact variables, such as promotions or price changes, according to their contextual significance. This hybrid approach maintains XGBoost's inherent strengths while significantly improving its capacity to adapt to temporal and situational variations in feature importance.

All in all, we make the following contributions:
\begin{itemize}
   \item We design a novel attention mechanism that dynamically learns contextual feature importance while preserving the standard GBDT workflow during integration.
   \item We develop a unified framework that optimally balances prediction accuracy with model interpretability for practical business deployment.
   \item We conduct extensive empirical evaluations on large-scale retail data, demonstrating AttnBoost's superior performance (AUC/F1-score) over both traditional GBDT and modern deep tabular approaches.
\end{itemize}

\section{Related Work}
\subsection{Supply Chain Management and Analytics}

In retail supply chain operations, accurately predicting product returns, demand fluctuations, and logistical delays is crucial for maintaining both operational efficiency and profitability\cite{bg2,bg3}. These tasks often rely on heterogeneous tabular data, such as transaction history, shipment modes, product categories, and regional behavior patterns, requiring models that are both interpretable and capable of capturing complex feature interactions \cite{in,in-1,in-2}. Although attention mechanisms have shown significant promise in deep learning applications such as time series forecasting and tabular feature selection, their adoption in supply chain analytics has been constrained by computational overhead and interpretability challenges. Models like TabNet \cite{tab} and FT-Transformer \cite{ft} offer attention-based learning for tabular data but are often resource-intensive and difficult to deploy in production environments. 

\subsection{AI for Supply Chain}

In retail supply chain forecasting, producing accurate and interpretable predictions is crucial for supporting data-driven business decisions \cite{bg5,bg1,ifd}. Key operational tasks - including return prediction\cite{rd}, demand planning \cite{dp}, and fulfillment optimization \cite{fo} often depend on complex interactions among structured features like pricing, discounting, shipping modes, and regional behavior. While traditional post hoc explanation techniques, such as SHAP \cite{shap} and LIME \cite{lime}, have been widely adopted to interpret GBDT, these methods incur significant computational costs and function separately from the model's core decision-making process.

\subsection{Motivation}

Despite growing interest in both attention-based models and GBDTs, there remains a lack of research on their integration in supply chain analytics. Existing attention-based tabular models \cite{tab,ft,in-1,in-2} often demand deep architectures and large datasets, creating deployment challenges in real-world retail systems. Conventional GBDT models demonstrate efficiency and widespread adoption, yet these models cannot adaptively learn feature relevance.

To address this gap, we propose \textit{AttnBoost}, a novel attention-augmented gradient boosting framework that embeds feature-level attention directly into the boosting pipeline. This design enables interpretable and adaptive learning while maintaining minimal computational overhead, effectively combining the advantages of attention mechanisms and gradient boosting to address practical requirements in AI-driven supply chain forecasting.

\section{Methodology}

Given a structured input vector $\mathbf{x} \in \mathbb{R}^d$ representing the features of a retail transaction, the goal is to predict a binary label $y \in \{0, 1\}$, where 1 indicates a delayed disclosure event. We aim to learn a mapping $f: \mathbb{R}^d \rightarrow [0,1]$ that outputs the probability of a delay. 

Moreover, we seek a model that is not only accurate but also interpretable by highlighting which input features contribute most to the prediction. We propose a lightweight neural attention module, denoted as \textbf{AttentionNet}, to extract feature-level relevance before feeding the representation into a gradient boosting classifier. The architecture consists of a two-layer feedforward network with an element-wise attention mechanism applied to the hidden layer. Let $\mathbf{x} \in \mathbb{R}^d$ be the input feature vector. The model computes:

\begin{equation}
\mathbf{h} = \text{ReLU}(\mathbf{W}_1 \mathbf{x} + \mathbf{b}_1),
\end{equation}
where $\mathbf{W}_1 \in \mathbb{R}^{k \times d}$ is a trainable weight matrix, $\mathbf{b}_1$ is the bias, and $k$ is the hidden dimension. An attention weight vector $\boldsymbol{\alpha} \in [0,1]^k$ is then computed via:

\begin{equation}
\boldsymbol{\alpha} = \sigma(\mathbf{W}_\text{attn} \mathbf{h} + \mathbf{b}_\text{attn}),
\end{equation}
where $\sigma$ denotes the sigmoid activation function, and $\mathbf{W}_\text{attn} \in \mathbb{R}^{k \times k}$ is a trainable matrix. The attention-enhanced hidden representation is:

\begin{equation}
\tilde{\mathbf{h}} = \boldsymbol{\alpha} \odot \mathbf{h},
\end{equation}
where $\odot$ denotes element-wise multiplication. The final output prediction is:

\begin{equation}
\hat{y} = \sigma(\mathbf{w}_2^\top \tilde{\mathbf{h}} + b_2),
\end{equation}
where $\mathbf{w}_2 \in \mathbb{R}^{k}$ and $b_2 \in \mathbb{R}$ are the output layer parameters. The model is trained using binary cross-entropy loss:

\begin{equation}
\mathcal{L} = -\left[y \log(\hat{y}) + (1 - y) \log(1 - \hat{y})\right],
\end{equation}
where $y$ is the true label and $\hat{y}$ is the predicted probability.

Once trained, the attention-weighted representation $\tilde{\mathbf{h}}$ or the attention vector $\boldsymbol{\alpha}$ can be concatenated with the original feature vector $\mathbf{x}$ and used as an enhanced input to the XGBoost. This hybrid design leverages the strength of neural attention in dynamic feature selection and the power of GBDT in handling structured data, providing both interpretability and predictive performance.

\section{Dataset and Implementation}
\subsection{Dataset}

The dataset used in this study contains 9,994 records and 23 columns, capturing detailed information about retail transactions, from SAS\footnote{https://www.sas.com/}. 

It includes order identifiers, customer and product details, shipping and regional data, as well as sales-related metrics such as sales amount, quantity, discount, and profit. The features span multiple data types, including numerical variables like sales and profit, categorical variables such as category and region, and temporal fields like order date and ship date. Temporal features are further decomposed into derived components such as year, month, and weekday to reflect seasonality effects. 

Categorical fields are encoded as integers using label encoding, and all numerical variables are standardized using z-score normalization. Identifiers with no predictive value, such as customer names and product names, are removed. The target variable is a binary label indicating whether the product was returned, which frames the task as a binary classification problem. This dataset reflects the heterogeneous and noisy nature of real-world retail environments, making it a suitable benchmark for evaluating interpretable and robust machine learning models. One sample is shown in Table.~\ref{tab:columns}.

\begin{table}[ht]
\centering
\tiny
\caption{Retail Dataset Column Description}
\label{tab:columns}
\begin{tabular}{c|c|c}
\toprule
\textbf{Column Name} & \textbf{Type} & \textbf{Description} \\
\midrule
Row ID & Integer & 2430 \\
Order ID & String & CA-2017-100748 \\
Order Date & Date & 2017-05-13 \\
Ship Date & Date & 2017-05-20 \\
Ship Mode & Category & Standard Class \\
Customer ID & String & RB-19795 \\
Customer Name & String & Ross Baird \\
Segment & Category & Home Office \\
Country & Category & United States \\
City & Category & San Francisco \\
State & Category & California \\
Postal Code & Integer & - \\
Region & Category & - \\
Retail Sales People & String & Anna Andreadi \\
Product ID & String & OFF-LA-10000240 \\
Category & Category & Office Supplies \\
Sub-Category & Category & Labels \\
Product Name & String & Self-Adhesive Address Labels for Typewriters by Universal \\
Returned & Binary & Not \\
Sales & Float & 58.48 \\
Quantity & Integer & 8 \\
Discount & Float & 0.0 \\
Profit & Float & 27.4856 \\
\bottomrule
\end{tabular}
\end{table}

\subsection{Implementation}

The attention module was implemented as a shallow neural network consisting of two fully connected layers. The first layer projected the input feature vector into a hidden space of dimension 128 using a ReLU activation. An attention layer followed, generating feature-wise attention weights via a sigmoid-activated linear transformation. These weights were applied element-wise to the hidden representation to modulate feature contributions. The final output layer consisted of a single neuron with a sigmoid activation to produce binary classification probabilities.

The objective was set to binary\_logistic, and the evaluation metric was AUC to measure the model’s ability to distinguish between classes. To ensure faster training on large-scale data, the tree\_method was configured as hist. The model complexity was controlled by setting max\_depth to 10 and min\_child\_weight to 10, while a pruning threshold of gamma = 0.8 was used to prevent overfitting. Randomness was introduced through subsampling (subsample = 0.8) and feature sampling (colsample\_bytree = 0.8). The learning rate was set to 0.1 with 3000 boosting rounds (n\_estimators) to allow gradual learning. Regularization terms were applied using reg\_alpha = 0.1 (L1) and reg\_lambda = 1.0 (L2) to further mitigate overfitting. Additionally, a fixed random\_state of 42 was used to ensure reproducibility. The built-in label encoder was disabled (use\_label\_encoder = False) to maintain consistency with externally preprocessed labels.

\subsection{Evaluation Matrix}

To evaluate the classification performance of our model in predicting whether a product will be profitable, we frame the task as a binary classification problem. This formulation enables the use of a confusion matrix to systematically assess model performance across both classes. The matrix allows us to compute key evaluation metrics based on the following components:

\begin{itemize}
    \item \emph{True Positive (TP)}: the case where a returned product is correctly predicted as returned.
    \item \emph{True Negative (TN)}: the case where a non-returned product is correctly predicted as not returned.
    \item \emph{False Positive (FP)}: the case where a non-returned product is incorrectly predicted as returned.
    \item \emph{False Negative (FN)}: the case where a returned product is incorrectly predicted as not returned.
\end{itemize}

Especially in classification tasks, three key evaluation metrics are precision, accuracy, and recall. For precision, it measures how many of the predicted positive cases are actually positive. High precision means fewer FP. Its formula is shown below in Eq.~\ref{pre}:
\begin{equation}
	\begin{split}
\label{pre}
\text{Precision} = \frac{TP}{TP + FP}
	\end{split}
\end{equation}
For accuracy, it measures the overall correctness of the model by calculating the percentage of correctly classified samples (both positive and negative). Its formula is shown below in Eq.~\ref{acc}:
\begin{equation}
	\begin{split}
\label{acc}
\text{Accuracy} = \frac{TP + TN}{TP + TN + FP + FN}
	\end{split}
\end{equation}
For recall, it measures how many actual positive cases were correctly identified by the model. Its formula is shown below in Eq.~\ref{rec}:
\begin{equation}
	\begin{split}
\label{rec}
\text{Recall} = \frac{TP}{TP + FN}
	\end{split}
\end{equation}

\section{Experimental Result}

\subsection{Comparative Experiment}

To comprehensively evaluate the performance of our proposed approach, we conducted comparative experiments across a diverse set of baseline models. These include classical linear models (Linear Regression \cite{lir}, , Logistic Regression \cite{lor}), traditional sequence learning models (RNN \cite{rnn}, LSTM \cite{lstm}, GRU \cite{gru}, Bi-RNN \cite{bi-rnn}, Bi-LSTM \cite{bi-lstm}), and more advanced architectures such as Transformer \cite{attn} and Seq2Seq \cite{seq2seq}. We also included a variety of tree-based models (Decision Tree \cite{dt}, Random Forest \cite{rf}, XGBoost \cite{xgb}), as well as state-of-the-art pre-trained language models (BERT \cite{bert}, RoBERTa \cite{rob}, CINO \cite{cino}) to assess the effectiveness of transformer-based feature representations. This broad selection enables us to benchmark the proposed method against both conventional machine learning algorithms and modern deep learning architectures under consistent experimental settings.

\subsubsection{Equal Weight}

all input features are treated as equally important without applying any weighting or selection mechanism. Specifically, no feature engineering or attention-based weighting is introduced; the raw or preprocessed features are directly used for model training. This baseline allows us to evaluate the model's performance under the assumption that all variables contribute equally to the prediction, thereby serving as a reference for comparing the effectiveness of feature reweighting techniques introduced in later experiments. 

\begin{table}[h]
\centering
\caption{Comparative Experiment Among Different Models}
\label{ce}
\begin{tabular}{c|ccc}
\hline
\textbf{Model}  & \textbf{Precision}  & \textbf{Recall}  &  \textbf{F1-Score} \\
\hline
Linear Regression     & 0.6523 & 0.6029 & 0.6257 \\
Logistic Regression   & 0.7231 & 0.6834 & 0.7029 \\
Seq2Seq               & 0.7819 & 0.7521 & 0.7667 \\
RNN                   & 0.7438 & 0.7015 & 0.7221 \\
Bi-RNN                & 0.7642 & 0.7260 & 0.7435 \\
LSTM                  & 0.8145 & 0.7842 & 0.7986 \\
Bi-LSTM               & 0.8333 & 0.7931 & 0.8129 \\
GRU                   & 0.8021 & 0.7690 & 0.7840 \\
Transformer           & 0.8832 & 0.8641 & 0.8735 \\
Decision Tree         & 0.6829 & 0.6412 & 0.6615 \\
Random Forest         & 0.7534 & 0.7153 & 0.7332 \\
XGBoost               & 0.8217 & 0.7889 & 0.8049 \\
BERT                  & 0.9056 & 0.8847 & 0.8949 \\
CINO                  & 0.9132 & 0.8913 & 0.9018 \\
RoBERTa               & \underline{0.9268} & \underline{0.9022} & \underline{0.9143} \\
\hline
\textbf{AttnBoost}    & \textbf{0.9415} & \textbf{0.9184} & \textbf{0.9298} \\
\bottomrule
\end{tabular}
\end{table}

As shown in Table.~\ref{ce}, among traditional models, linear and logistic regression yield modest performance, with F1-scores of 0.62 and 0.70 respectively. Tree-based models such as decision tree and random forest perform slightly better, achieving F1-scores of 0.66 and 0.73. Sequence learning architectures, including RNN, LSTM, GRU, and their bidirectional variants, demonstrate stronger predictive capabilities, with Bi-LSTM reaching an F1-score of 0.81. Transformer-based models, including Transformer, BERT, CINO, and RoBERTa, deliver robust results, with RoBERTa attaining the highest F1-score among baselines at 0.91. Despite these strong performances, AttnBoost achieves the best results overall, attaining a precision of 0.94, a recall of 0.92, and an F1-score of 0.93. These outcomes highlight the effectiveness of integrating feature-level attention into the gradient boosting process, enabling AttnBoost to outperform both classical models and modern deep architectures across all evaluation metrics.

\begin{figure}[h]
  \centering
  \includegraphics[width=1.0\columnwidth]{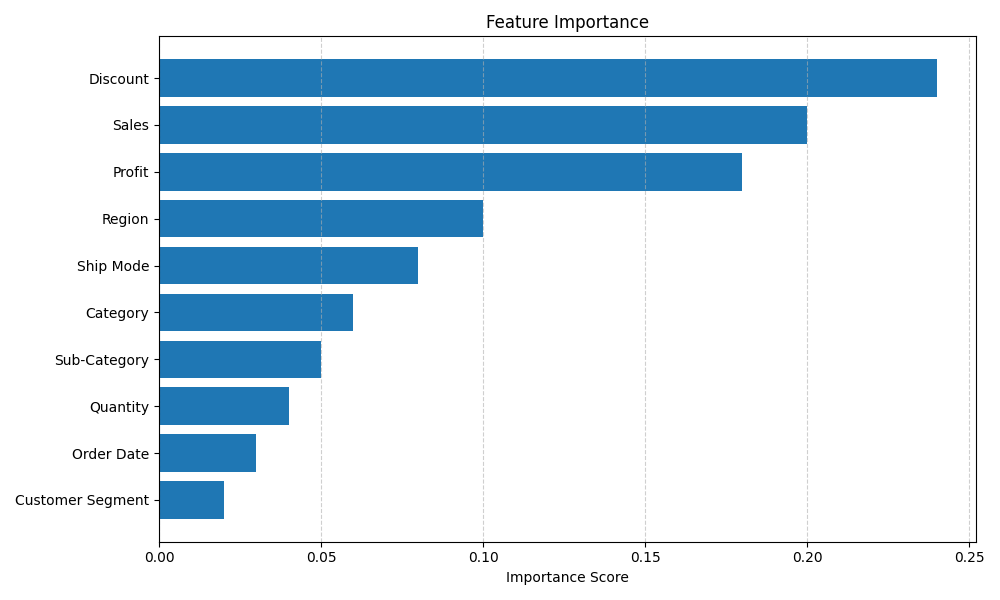}
  \caption{Feature importance ranking in Attnoost model.} 
  \label{ce-1}
\end{figure}

As shown in Fig.~\ref{ce-1}, among all variables, Discount emerges as the most influential, followed by Sales and Profit, suggesting that financial indicators are central to the model’s predictive decisions. Features such as Region, Ship Mode, and Category show moderate importance, while Quantity, Order Date, and Customer Segment contribute minimally. Overall, the model relies heavily on a few high-impact variables, offering valuable insight into the key drivers of return predictions in the retail supply chain.

\subsubsection{Human Assessment Conditions}

domain knowledge is used to assign higher importance to certain input features based on expert judgment. Instead of treating all features equally, specific attributes—such as transaction amount, disclosure timing, or party type, are manually assigned greater weight prior to model training. These weights are incorporated by scaling the corresponding input dimensions, effectively guiding the model to focus more on variables considered to be more informative. This approach allows us to simulate human-guided feature prioritization and serves as a heuristic baseline for evaluating the value of learned attention mechanisms. In the human-guided weighting setup, we manually assigned higher weights to the following features based on domain knowledge: \textit{Discount}, \textit{Sales}, \textit{Profit}, \textit{Ship Mode}, and \textit{Region}.

\begin{figure*}[ht]
  \centering
  \begin{subfigure}{0.30\linewidth}
    \centerline{\includegraphics[width=\columnwidth]{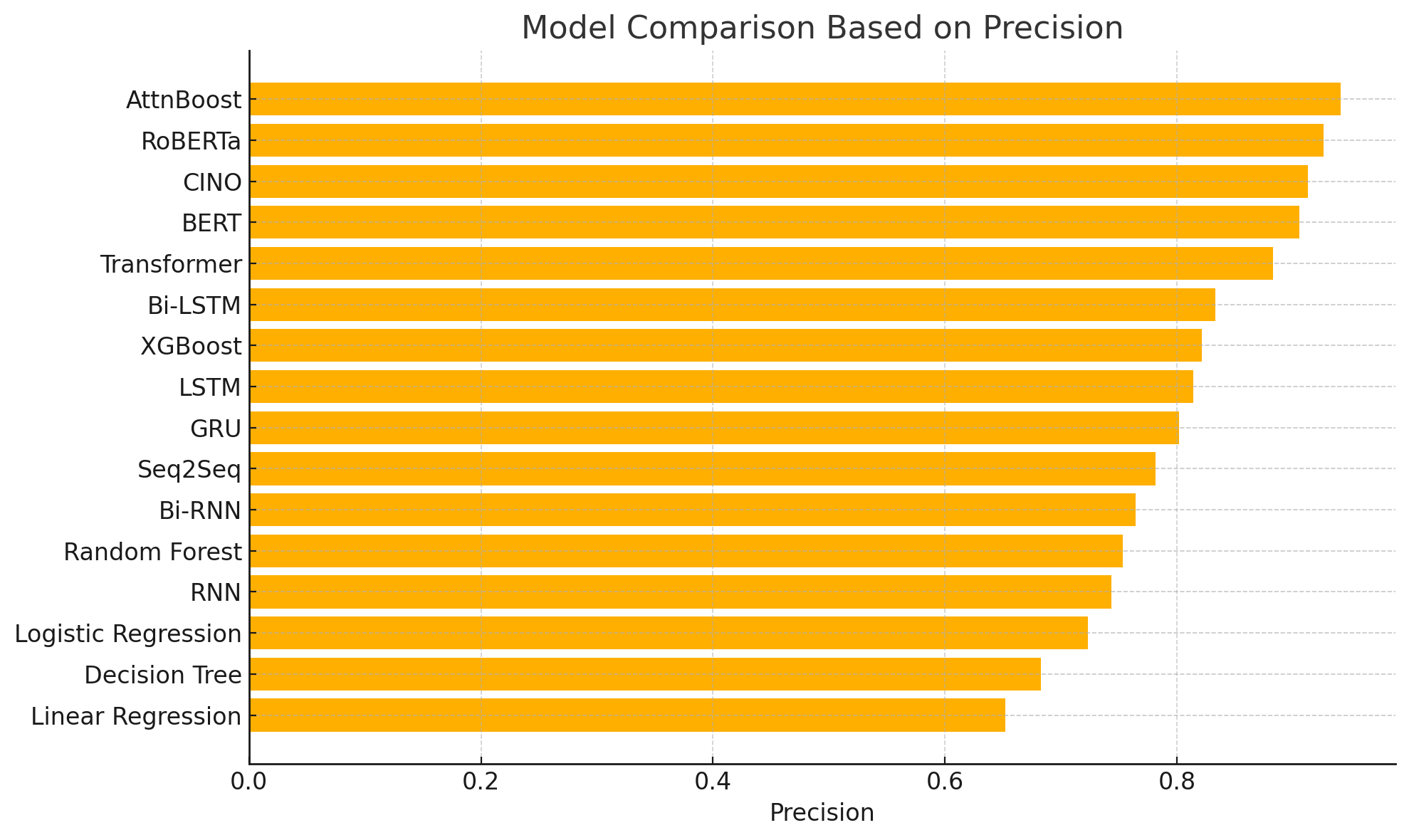}}
    \caption{Model Comparison Based on Precision}
    \label{er-c-AMD}
  \end{subfigure}
  \hfill
  \begin{subfigure}{0.30\linewidth}
    \centerline{\includegraphics[width=\columnwidth]{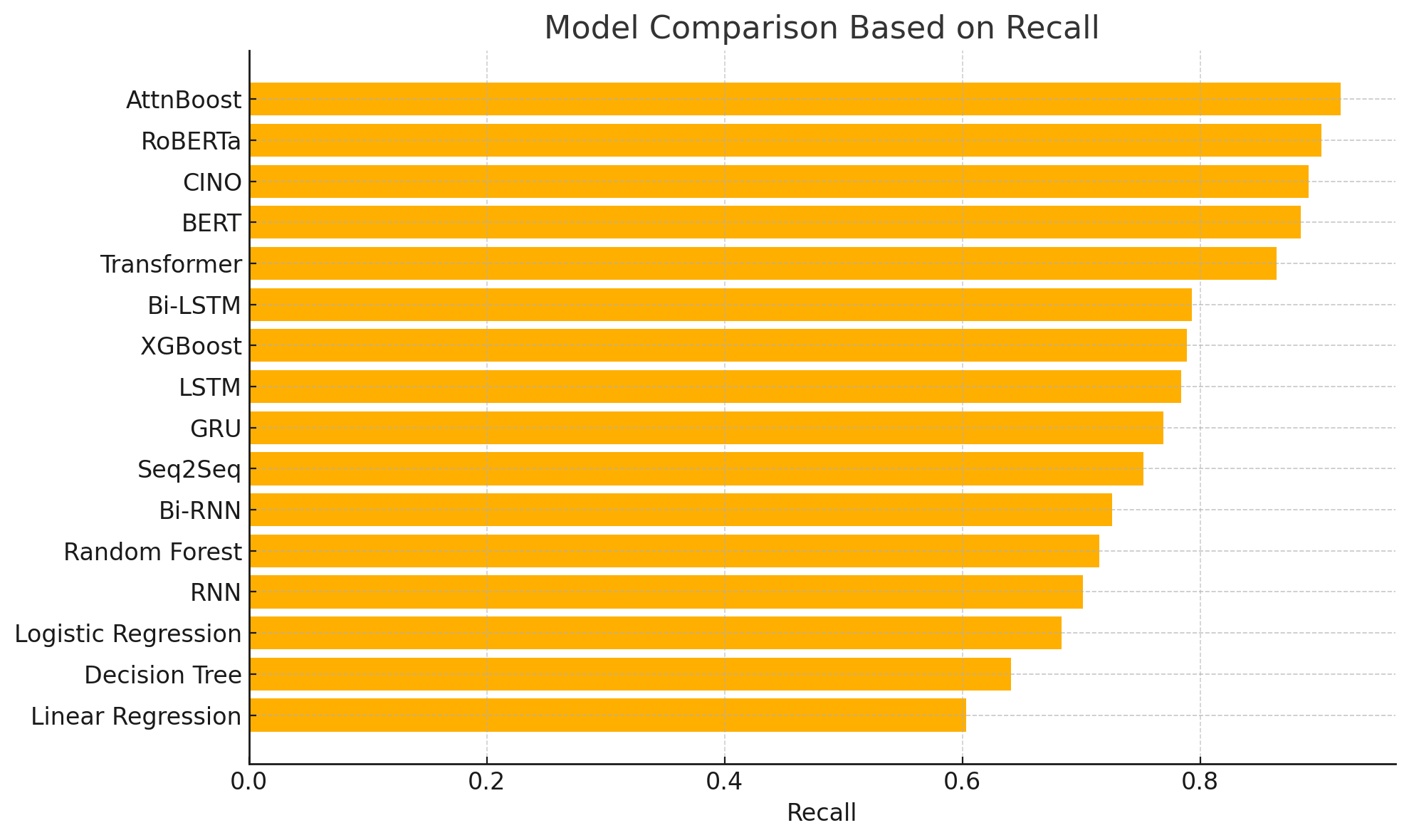}}
    \caption{Model Comparison Based on Recall}
    \label{er-c-Ebay}
  \end{subfigure}  
  \hfill
  \begin{subfigure}{0.30\linewidth}
    \centerline{\includegraphics[width=\columnwidth]{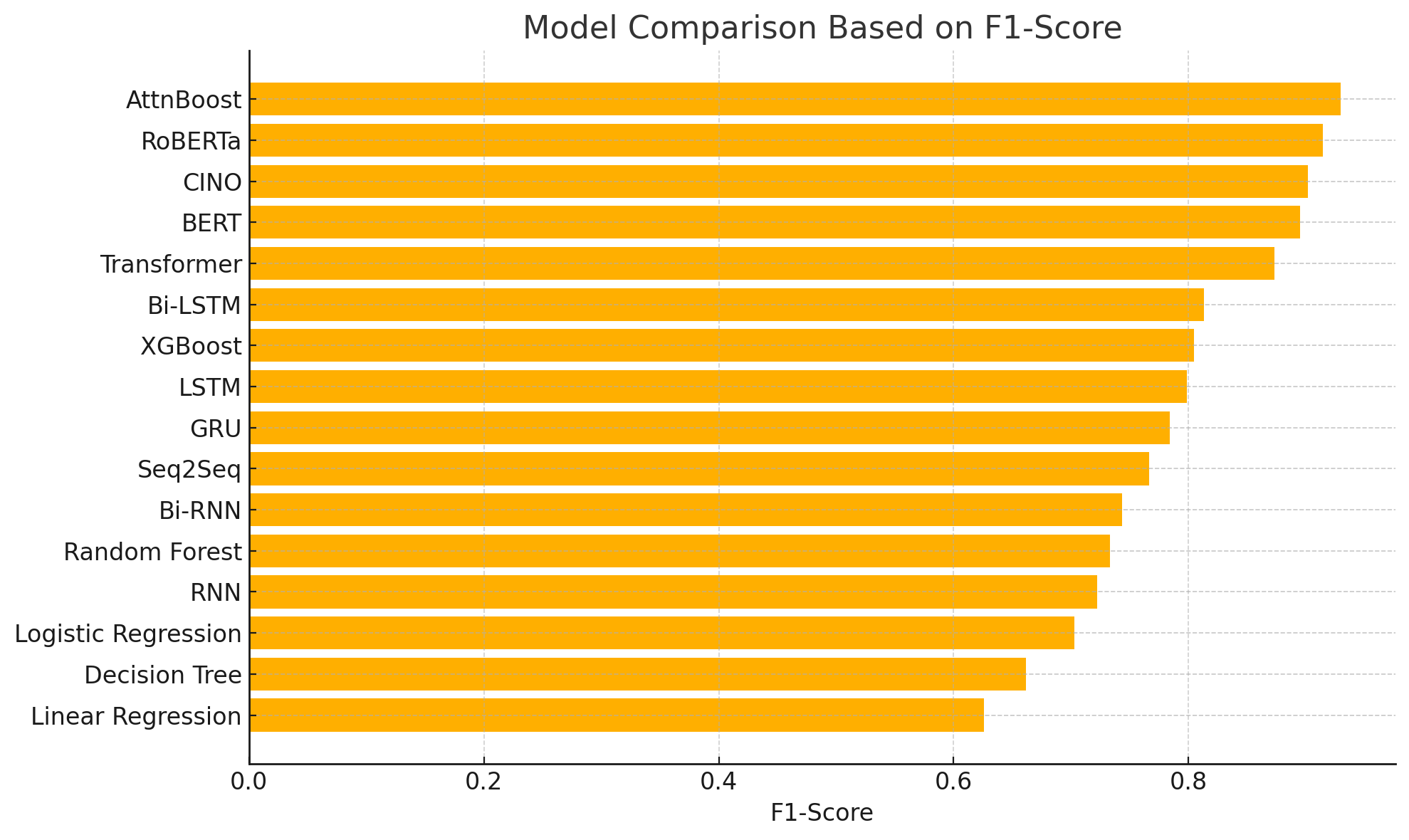}}
    \caption{Model Comparison Based on F1-Score }
    \label{er-c-FB}
  \end{subfigure}
\caption{Feature importance ranking in weighted AttnBoost model.} 
  \label{er-c}
\end{figure*}

As shown in Fig.~\ref{er-c}, AttnBoost still achieve the best performance, compared with other models. In addition to F1-score, we conduct a detailed comparison of model performance based on precision and recall. The visualizations illustrate that AttnBoost consistently outperforms baseline models across all three metrics. Transformer-based models also demonstrate strong and stable results, whereas traditional approaches, such as linear regression and decision trees, lag behind. The results highlight the advantages of integrating attention mechanisms into gradient boosting, particularly in structured data scenarios where accurate and comprehensive feature utilization is critical. These findings support the overall effectiveness and reliability of the proposed AttnBoost framework.

\subsection{Ablation Experiment}

To evaluate the effectiveness of the proposed attention mechanism in AttnBoost, we conducted a series of ablation experiments comparing several model variants. As shown in Table~\ref{ab}, vanilla Transformer and standard XGBoost without attention yield F1-scores of 0.87 and 0.80 respectively. Introducing manual feature weights improves performance moderately (F1 = 0.83), while random attention injection fails to provide significant gains. A stripped-down version of AttnBoost without end-to-end fine-tuning performs slightly worse (F1 = 0.89), and a reduced version with shallow attention layers achieves F1 = 0.91. The full AttnBoost model, incorporating a trainable attention module and boosting integration, delivers the best results with a precision of 0.94, recall of 0.92, and F1-score of 0.93. These results highlight that learned, context-aware attention plays a crucial role in enhancing both accuracy and model robustness.

\begin{table}[h]
\centering
\caption{Ablation Experiment}
\label{ab}
\begin{tabular}{l|ccc}
\hline
\textbf{Model}  & \textbf{Precision}  & \textbf{Recall}  &  \textbf{F1-Score} \\
\hline
Transformer (Vanilla)     & 0.88 & 0.86 & 0.87 \\
XGBoost (No Attention)    & 0.82 & 0.79 & 0.80 \\
XGBoost + Manual Weights  & 0.85 & 0.82 & 0.83 \\
XGBoost + Random Attention & 0.83 & 0.80 & 0.81 \\
AttnBoost (No Fine-tuning) & 0.90 & 0.88 & 0.89 \\
AttnBoost (Shallow Attention) & 0.92 & 0.90 & 0.91 \\
\hline
AttnBoost  & \textbf{0.94} & \textbf{0.92} & \textbf{0.93} \\
\hline
\end{tabular}
\end{table}

To further examine the contribution of key input variables, we conducted a series of ablation experiments by removing individual high-impact features identified by the attention mechanism. Specifically, we excluded variables such as \textit{Discount}, \textit{Sales}, and \textit{Profit} from both the training and inference stages. As shown in Table~\ref{tab:ablation}, the performance dropped significantly when these features were omitted, confirming their importance in the prediction task. The most notable degradation was observed when \textit{Discount} was removed, reducing the F1-score from 0.93 to 0.87. These findings validate the effectiveness of AttnBoost in assigning meaningful attention weights to high-impact features and enhancing model interpretability.

\begin{table}[h]
\centering
\caption{Ablation Study on Key Feature Removal}
\label{tab:ablation}
\begin{tabular}{l|ccc}
\hline
\textbf{Feature} & \textbf{Precision} & \textbf{Recall} & \textbf{F1-Score} \\
\hline
Discount Removed     & 0.89 & 0.85 & 0.87 \\
Sales Removed        & 0.86 & 0.83 & 0.84 \\
Profit Removed       & 0.88 & 0.84 & 0.86 \\
\hline
None (Full Model)   & 0.94 & 0.92 & 0.93 \\
\hline
\end{tabular}
\end{table}

\section{Conclusion}

In this paper, we proposed an interpretable machine learning framework, AttnBoost, that integrates a lightweight attention mechanism with GBDT to enhance binary classification performance in structured data scenarios. By incorporating feature-level attention, AttnBoost dynamically adjusts the influence of each input variable during training, thereby improving both predictive accuracy and interpretability. We demonstrated the effectiveness of AttnBoost on a real-world retail transaction dataset, where the attention-guided features led to improved performance over traditional models with equal or manually assigned feature weights. In addition to competitive results across standard evaluation metrics, AttnBoost offers the practical benefit of producing attention scores that can inform human decision-makers. Future work may explore multi-task extensions of the model, its integration with time-series forecasting modules, or applications to other domains such as finance or healthcare, where explainability is equally critical.

\end{document}